\documentclass[letterpaper, 10 pt, conference]{ieeeconf}
\IEEEoverridecommandlockouts
\usepackage[linesnumbered,ruled,noend,algo2e]{algorithm2e}
\usepackage{algorithm}
\usepackage{subcaption}
\usepackage{amsmath}
\usepackage{mathabx}
\usepackage[bookmarks=true]{hyperref}
\usepackage{graphicx}
\usepackage{times}
\usepackage{underscore}
\usepackage{moreverb}
\usepackage[dvipsnames]{xcolor}
\usepackage{url}
\usepackage{multicol}
\usepackage[utf8]{inputenc}
\usepackage{graphics}
\usepackage{bm}
\usepackage[noend]{algpseudocode}
\usepackage{lipsum}
\usepackage{dblfloatfix}
\usepackage[english]{babel}
\usepackage{blindtext}
\usepackage{mathtools}
\usepackage{booktabs}
\usepackage{float}
\usepackage{cite}

\renewcommand\vec{\mathbf}
\SetKwRepeat{Do}{do}{while}

\begin{document}

\title{The Busboy Problem: Efficient Tableware Decluttering 

Using Consolidation and Multi-Object Grasps }

\author{Kishore Srinivas$^{1}$, Shreya Ganti$^{1}$, Rishi Parikh$^1$, Ayah Ahmad$^1$, \\ Wisdom Agboh$^{1,2}$, 
Mehmet Dogar$^{2}$, Ken Goldberg$^{1}$
\thanks{$^{1}$The AUTOLab at UC Berkeley (automation.berkeley.edu).}
\thanks{$^{2}$University of Leeds, UK.}
}
\maketitle

\begin{abstract}
We present the ``Busboy Problem": automating an efficient decluttering of cups, bowls, and silverware from a planar surface. As grasping and transporting individual items is highly inefficient, we propose policies to generate grasps for multiple items. We introduce the metric of Objects per Trip (OpT) carried by the robot to the collection bin to analyze the improvement seen as a result of our policies. In physical experiments with singulated items, we find that consolidation and multi-object grasps resulted in an 1.8x improvement in OpT, compared to methods without multi-object grasps. See https://sites.google.com/berkeley.edu/busboyproblem for code and supplemental materials.
\end{abstract}
    \vspace{-1mm}
\section{Introduction}
The post-meal task of clearing a dining table, commonly referred to as ``bussing,'' requires moving cups, bowls, and utensils that are dispersed across the surface into a bin or tray to be cleaned in the kitchen. This is a common task that occurs after any event involving food service and dish collection, from daily household meals to casual picnics to formal cocktail parties and dinners. Automating this tedious and repetitive task could reduce fatigue and busy work for the skilled waiters who typically perform it.

We define the ``Busboy Problem" as the efficient transfer of cups, bowls, and utensils (collectively called tableware) from the table into a designated collection bin while minimizing the time required for completion. This is an interesting problem for automation because the tableware are of varying shape, requiring low-level planning to execute grasps and high-level planning to consolidate tableware for efficient transport. Even small inaccuracies can lead to toppling or dropping delicate and expensive tableware, so the system must be extremely reliable.
    \begin{figure}[ht]
        \includegraphics[width=0.95\linewidth]{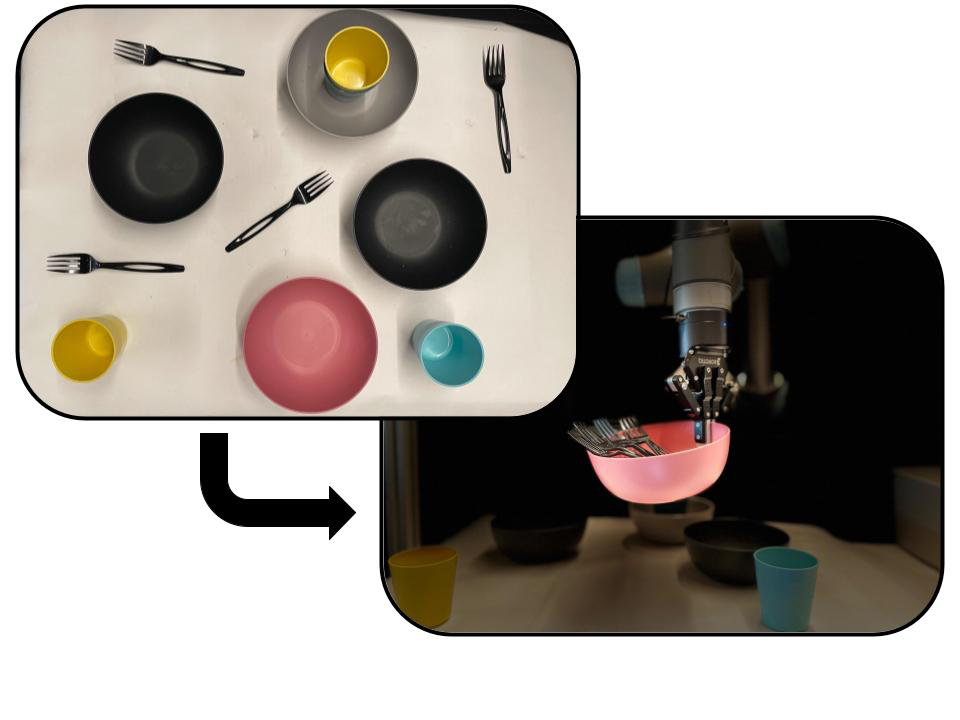}
        \vspace{-4mm}
        \caption{The Busboy Problem. We present a combination of the robust action primitives to declutter a workspace of cups, bowls, and utensils.}
        \label{fig:opt}
    \end{figure}
\begin{figure*}
\begin{center}
    \includegraphics[width=\textwidth]{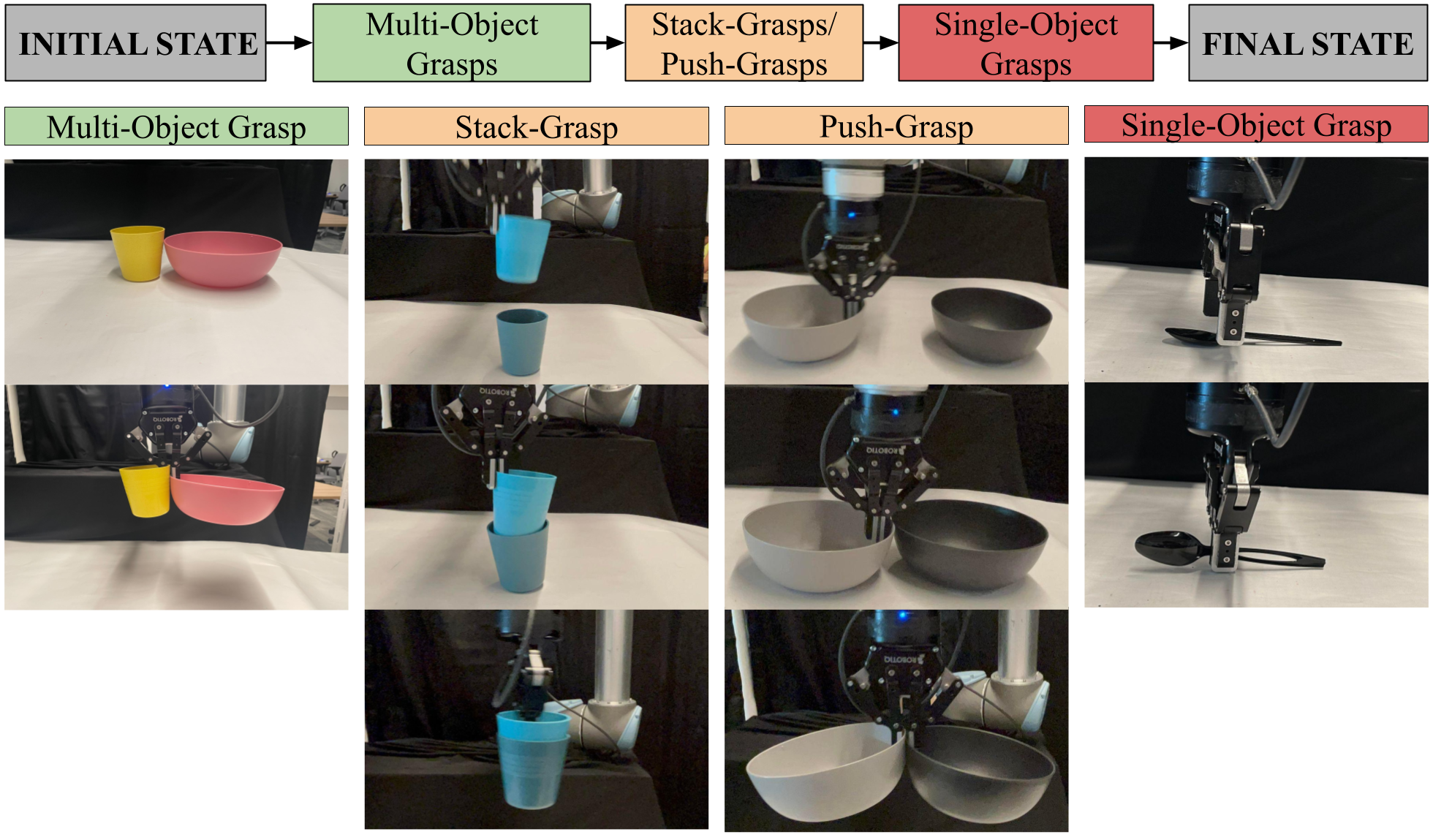}
    \caption{We employ the following robust action primitives to declutter a workspace through consolidation and multli-object grasps: single-object grasps, multi-object grasps, push-grasps, and stack-grasps.}
    \label{fig:splash}
    \vspace{-5mm}
\end{center}
\end{figure*}

Previous work in multi-object grasping, object manipulation, and grasp candidate generation highlight the efficiency of grasping pre-stacked objects as well as objects manually oriented for multi-object grasps \cite{DBLP:journals/corr/abs-2110-06192, yamada2009grasp}. Whereas these works explore situations with objects are already positioned for said grasps, our work investigates methods of stacking and clustering objects into these favorable positions for multi-object grasps.

In this paper, we present a framework and algorithms for the Busboy Problem. We consider a scenario where multiple items are placed on a work surface (see Fig. \ref{fig:splash}), under an RGBD camera. We use the concept of multi-object grasping, which enables the robot to move multiple items simultaneously, thus reducing the number of pick-and-place actions needed. 

This paper makes the following contributions:
\begin{enumerate}
    \item {Formulation of the Busboy Problem.}
    \item {Action primitives for rearranging and grasping cups, bowls, and utensils.}
    \item {Two algorithms that leverage consolidation and multi-object grasps.}
    \item{Experimental results indicating a 1.8x improvement in OpT.}
\end{enumerate}
    \vspace{-3mm}
\section{Related work}
\subsection{Multi Object Grasping}  
Prior work on multi-object grasping includes different grasping techniques to facilitate multi-object grasps \cite{9812388}, detecting the number of objects in a grasp \cite{9636777}, decluttering surfaces \cite{Agboh-ISRR-2022}, and multi-object grasping to place objects in virtual reality \cite{app12094193}. Yamada et al. considered the simplified multi-object grasping problem, where the objects are already in a configuration where they can be grasped at once \cite{yamada2009grasp}. Agboh et. al. \cite{agboh2022learning} 
showed that friction can increase picks per hour for convex polygonal objects. 

Some prior work has focused on the design of grippers for multi-object grasping. Jiang et. al. \cite{jiang2023multipleobject} proposed a vaccum gripper with multiple suction cups, while Nguyen et. al. \cite{Nguyen-MOG} proposed a soft gripper based on elastic wires for multi-object grasping. 

Object stacking \cite{DBLP:journals/corr/abs-2110-06192, https://doi.org/10.48550/arxiv.2207.02347, doi:10.1177/0278364912438781} has the potential to improve the number of objects per trip. We take inspiration from these works to include a stacking primitive. 
    \vspace{-1.5mm}
\subsection{Pulling}
Prior work by Berretty et al. has examined the use of inside-out pulling to orient convex polygonal parts \cite{Berretty2001OrientingPB}. We utilize a similar technique for circular cups and bowls. Furthermore, a planner for ensuring convergence to the final pose of pulling trajectories is proposed by Huang et al. \cite{Huang2017ExactBO}, where they examine the motion of planar objects undergoing quasi-static movement.
    \vspace{-1mm}
\subsection{Grasp Candidates}
Satish et al. discuss using a synthetic data sampling distribution that combines grasps sampled from the policy action set with guiding samples from a robust grasping supervisor to construct grasp candidates \cite{Satish-RAL-2019}.
Additionally, Mahler et al. \cite{8360035} discuss the use of energy-bounded caging to evaluate grasp candidates. They efficiently compute candidate rigid configurations of obstacles that form energy-bounded cages of an object, where the generated push-grasps are robust to perturbations.
Mousavian et al. describe the process of using a variational autoencoder to generate grasps by mapping the partial point cloud of an observed object to a diverse set of grasps for the object \cite{https://doi.org/10.48550/arxiv.1905.10520}. 
Because of the relative simplicity of our setup, we found that an analytical approach to constructing grasp candidates is sufficient. In the case of bowls and cups, we sample a random point uniformly on the rim and then orient the gripper perpendicular to the tangent of the circle at that point. In the case of utensils, we identify the axis of the utensil, and pick the highest depth point along that line, with the gripper perpendicular to the axis.
    \vspace{1mm}
\subsection{Object Manipulation in Cluttered Environments}
    \vspace{-1mm}
Efficiently finding object manipulation plans in high-dimensional environments with a large number of objects is a challenging problem. Hasan et al. \cite{DBLP:journals/corr/abs-2002-12738} addressed this problem by identifying high-level manipulation plans in humans, and transferring these skills to robot planners. Other work by Tirumala et al. \cite{Tirumala-IROS-2022} used tactile sensing to singulate layers of cloth from a stack. 
Different from these works, our goal in the cluttered environment is to bring objects together, or stack them, to enable multi-object grasps.
\vspace{-1mm}
\section{The Busboy Problem} 
The Busboy Problem involves the task of decluttering a workspace containing cups, bowls, and utensils, with the objective of minimizing both the time and number of trips required for completion.
    \vspace{-1mm}
\subsection{Assumptions}
    \vspace{-1mm}
    In the initial configuration, a planar workspace is defined in a cartesian grid $(x, y)$ and has $n_c$ cups, $n_b$ bowls, and $n_u$ utensils scattered across its surface. 
    All items are assumed to be face up, visible by camera, and within a workspace defined by the constraints of the robot arm. These items may be initially stacked on top of one another or resting individually on the surface, and we assume that the initial state meets the following criteria:
    \begin{itemize}
        \item {All items are of known dimensions, and cups and bowls are circular when viewed from top-down. Cups have radius 4.5cm, bowls have radius 8.5cm, and utensils are at most 17cm $\times$ 1.8cm.}
        \item {Cups and bowls are upright, and utensils are laid flat on the surface.}
        \item {Any stacks that exist are stable, such that $r_0 \geq r_1 \geq ... \geq r_{s}$, where $r_0$ represents the radius of the vertically lowest item, and $r_s$ the highest one.}
        \item {Initially, no two items are touching (items are singulated).}
    \end{itemize}
\vspace{-2mm}
\subsection{State}
We use cups, bowls, and utensils (forks and spoons) as the tableware set - collectively called ``tableware" - in this work. 
Each cup and bowl has a position $[x, y]$, and each utensil has a position $[x, y]$ and orientation $\theta$. 

\section{Decluttering Tableware} 
    \subsection{Action primitives}
    We propose to use a combination of manipulation primitives to solve the Busboy Problem. We specifically propose to use single object grasps, multi-object grasps, pull-grasps, and stack-grasps to efficiently clear a work surface of items (Figure \ref{fig:splash}). 
    \begin{figure}
        \includegraphics[width=0.95\linewidth]{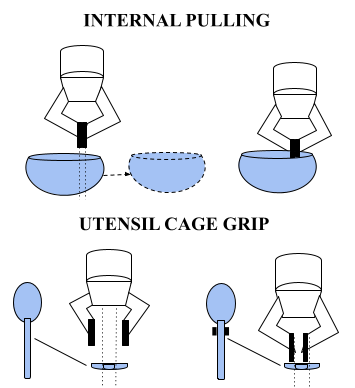}
        \caption{Bowls and cups are pulled with an internal pull, and utensils are grasped with a cage grip.}
        \label{fig:pulling}
    \end{figure}
    
    \subsubsection{Grasp} 
    We use both single and multi-object grasps in this work. Let $\vec{u}_{G}$ be the grasp to pickup objects --- single or multiple. We represent this action as:
        \vspace{-2mm}
    \begin{align}
        \vec{u}_{G} = [\vec{p}_{G}, \theta_{G}]
    \end{align}
    where $\vec{p}_{G} = [x_{G}, y_{G}, z_{G}]$ is the center point of the grasp, and $\theta_{G}$ is the grasp orientation. 
    
    \subsubsection{Pull-Grasp}
    A pull-grasp action involves two steps: a pull of one object to another, then a multi-object grasp of both objects. We represent a pull action as: 
        \vspace{-2mm}
    \begin{align}
        \vec{u}_{P} = [\vec{p}_{S}, \theta_{S}, \vec{p}_{E}, \theta_{E}]
    \end{align}
    where $\vec{p}_{S} = [x_{S}, y_{S}, z_{S}]$ is the pull start point, $\theta_{s}$ is the gripper orientation at the state point,  $\vec{p}_{E} = [x_{E}, y_{E}, z_{E}]$ is the pull end point, and $\theta_{E}$ is the gripper orientation at $\vec{p}_{E}$. For circular objects such as bowls and cups, the gripper pulls outwards from the center of the dish using an internal pull, and for utensils, the gripper cages the utensil around its center point while moving it (Figure \ref{fig:pulling}). Then, we denote a pull-grasp action as: 
        \vspace{-2mm}
    \begin{align}
     \vec{u_{PG}} = [\vec{u}_{P}, \vec{u}_{G}]
    \end{align}
    
    \subsubsection{Stack-Grasp}
    A stack-grasp action involves two steps: a stack of one object onto another, then a multi-object grasp of both objects. We represent a stack action as:
        \vspace{-5mm}

    \begin{align}
        \vec{u_{S}} = [\vec{u}_{G_i}, \vec{p}_{L}, \theta_{L}]
    \end{align}
    where $\vec{u}_{G_i}$ is a grasp on the lifted object, and $\vec{p}_{L} = [x_{L}, y_{L}, z_{L}]$ is the placement point on the stationary object, and $\theta_{L}$ is the gripper orientation at $\vec{p}_{L}$. Then, we denote a stack-grasp action as: 
        \vspace{-1mm}
    \begin{align}
     \vec{u_{SG}} = [\vec{u}_{S}, \vec{u}_{G}]
    \end{align}
\subsection{Determining allowable actions}
    \subsubsection{Grasp}
    A single-object grasp is always allowable. We can safely assume this since any dish or stack of items is already top-down graspable. When no other actions are allowed, the single-object grasp action is used as a default to clear the workspace. 
    
    A multi-object grasp is allowable when the grasp heights of both items are similar (within an adjustable threshold value) and if the lateral distance between the grasp points of both items is less than the width of the gripper. If the grasp heights of the items are significantly different, the gripper will have to either collide with the taller dish while attempting to grasp the shorter dish or grasp only the taller dish to avoid the collision, and either case results in a failure of grasping multiple items at once. Similarly, if the items are separated by more than the maximum inside width of the grippers, an attempt to grasp both at the same time will fail.
    \subsubsection{Pull}
    A pull of two items is allowable if a multi-object grasp can be executed on those items and if no other objects lie between the two items on the workspace. We disallow pull actions of items for which a multi-object grasp cannot be executed, since the pull becomes a wasted action. We also disallow pull actions of items with other objects between them to ensure that the intermediate objects are not displaced in a non-deterministic manner.
    \subsubsection{Stack}
    A stack of dish $d_a$ with radius $r_a$ onto dish $d_b$ with radius $r_b$ is allowable if $r_a \leq r_b$. This means that a cup can be stacked onto a bowl, but not vice versa, and that a utensil can be stacked onto any other dish, including another utensil. This is to ensure that the stack stability assumption present at the initial state remains valid after each action.

\vspace{-1mm}
\subsection{Robustness of action primitives}
    \begin{figure}
        \includegraphics[width=0.9\linewidth]{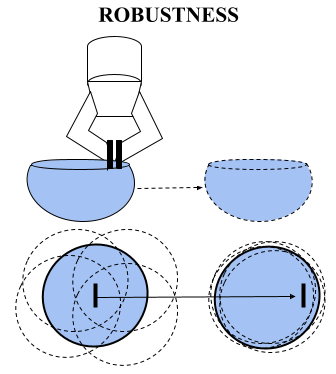}
        \caption{Uncertainty in a cup or bowl's location decreases once the arm has interacted with the dish. The position of the dish relative to the gripper's location is known with greater certainty after the action.}
        \label{fig:robustness}
    \end{figure}
    We present three primitives to robustly execute the above actions. This design makes the primitives more robust.
    \subsubsection{Grasp} When executing a grasp at location $x, y, z$, the robot will open its grippers centered around $x, y$, and then move down to the appropriate height, as measured by the depth sensor, before closing the gripper to grasp the object. The affordances granted by max gripper opening, gripper height, and gripper width mean that an off-center grasp point $x, y, z$ will still successfully complete the single-object or multi-object grasp of the object (Figure \ref{fig:robustness}).
    \subsubsection{Pull} For cups and bowls, the gripper pulls outwards from the center of the dish, contacting the inner surface of the dish (Figure \ref{fig:pulling}). This action is successful as both $r_b$ and $r_c$ are larger than the width of the gripper when closed. If the gripper is anywhere within the opening of the object, it will be able to move the target object to a specified location. For utensils, the gripper cages the utensil around its center point while moving it, preventing unwanted rotation and moving the utensil to its specified location. 
    \subsubsection{Stack} For bowls and cups, the top lip radius is larger than the radius of the base, giving the sides a taper. Because a dish $d_a$ is only stacked onto another dish $d_b$ of equal or larger size, the base radius of $d_a$ is guaranteed to be smaller than the top radius of $d_b$, allowing the tapered sides of the items to funnel $d_a$ into place even if there is slight error in the placement of the dish. Placing a utensil onto a bowl is extremely robust to error because of the relative radii of the items, and placing a utensil onto another utensil is robust due to the curvature of the utensils themselves which slide a misplaced utensil into place, making them naturally conducive to stacking.
    \vspace{-2mm}
\subsection{Policies}
\vspace{-1mm}
\subsubsection{Pull Policy}
    The pull policy combines Pull-Grasp and Grasp actions. From the initial scene, it checks if any multi-object grasps can be executed right away, and executes those first. Then, it runs the Pull-Grasp action for all remaining items, pulling together items that don't cause collisions and executing multi-object grasps to clear them from the workspace. If any items remain after all possible multi-object grasps are executed, those items are cleared with single-object Grasp actions. After each action, a new image of the workspace is taken and the state representation is updated to reflect the new state of the workspace, including any tableware that has been moved or left behind by the previous action. This policy is formalized in Algorithm \ref{alg:pull_policy}.
\subsubsection{Stack Policy} \label{stack_policy_subsection}
    The stack policy combines Stack-Grasp and Grasp actions. It repeatedly executes the Stack-Grasp action to clear the workspace, and if there are any remaining items they are cleared with single-object Grasp actions. It prioritizes stacking utensils onto bowls and transporting them to the bin, and then tries to stack the remaining dishes. Stacking utensils first is an efficient way to improve the number of OpT for this policy. The policy is formalized in Algorithm \ref{alg:stack_policy}.

    After utensils are cleared, the stacks created by this policy are limited to be a combination of at most 2 existing stacks (i.e. once a Stack action is executed, the next action is necessarily a Grasp on the resulting stack, not another Stack action onto that stack). This is because when 4 or more bowls or cups are stacked, the height difference between the lip of the top dish and the lip of the bottom dish exceeds the height of the gripper jaws, causing many attempted grasps to fail. By limiting stacks to at most 2 existing stacks, we significantly reduce the chances of creating a stack with more than 3 dishes.

\begin{algorithm}

\caption{Pull Policy
\\ \textbf{Input:} $D \gets [d_1, d_2, ... d_n]$ - initial distribution of $n$ items to be cleared.
\\ \textbf{Output:} Executes an action sequence at each iteration.
}\label{alg:pull_policy}

\While{$D \neq \emptyset$}{
    \If{MultiObjGraspPossible($d_i, d_j$) $\forall i, j \in$ len(D)}{
        Grasp($d_i, d_j$);\\
        D $\gets$ UpdateStateFromImage(); \\
        continue;
    }
    \If{PullGraspPossible($d_i, d_j$) $\forall i, j \in$ len(D)}{
        Pull($d_i, d_j$);\\
        Grasp($d_i, d_j$);\\
        D $\gets$ UpdateStateFromImage(); \\
        continue;\\
    }
    Grasp($d_i$) $\forall i \in$ len(D);\\
    D $\gets$ UpdateStateFromImage();
}
\end{algorithm}
\vspace{-2mm}
\begin{algorithm}

\caption{Stack Policy
\\ \textbf{Input:} $D \gets [d_1, d_2, ... d_n]$ - initial distribution of $n$ items to be cleared.
\\ \textbf{Output:} Executes an action sequence at each iteration.
}\label{alg:stack_policy}

\While{$D \neq \emptyset$}{
    \If{UtensilsAndBowlsRemaining()}{
        Stack($u, b$) $\forall u \in$ {utensils};\\
        Grasp($b$);\\
        D $\gets$ UpdateStateFromImage(); \\
        continue;
    }
    \If{StackGraspPossible($d_i, d_j$) $\forall i, j \in$ len(D)}{
        Stack($d_i, d_j$);\\
        Grasp($d_j$);\\
        D $\gets$ UpdateStateFromImage(); \\
        continue;\\
    }    
    Grasp($d_i$) $\forall i \in$ len(D);\\
    D $\gets$ UpdateStateFromImage(); 
}
\end{algorithm}

\vspace{-1mm}
\section{Experiments and Results} 
We evaluate through physical experiments the robustness of the pulling action primitive and then evaluate the pull and stack policies on a real-world table clearing task. 

\vspace{-2mm}
\subsection{Experimental Setup}
We use a UR5 robot arm with a Robotiq 2F-85 gripper and Intel RealSense 455D RGBD camera mounted 83cm above the workspace. The workspace is a flat 78cm x 61cm surface with 4 cups, 4 bowls, and 4 utensils, $n_b = n_c = n_u = 4$. In our experimental setup, we calculated a max gripper opening of $w = 8.5cm$, gripper height of $h = 4.5cm$, bowl radius $r_b = 8.5cm$, cup radius $r_c = 4.5cm$ and utensil width $r_u = 1.8cm$.

We identify and locate tableware on the workspace with a vision pipeline. Since the surface of the workspace is white, we use darker colored tableware to be easily visible. To locate cups and bowls, we first use edge detection, contour forming, and HoughCircles to identify circular shapes on the workspace, then filter these circles based on the known image radius of cups and bowls. We cluster these circles by their centers and remove circles that overlap beyond a specified threshold, allowing an unambiguous detection of cups and bowls. To locate utensils, we use edge detection and contour forming, and then filter out the contours that are too ``square", as determined by the aspect ratio of the identified contour. We draw an imaginary line through the lengthwise center of bounding rectangle of the contour, and sample depth values along that line; we use the highest depth point as the grasp point of the utensil to allow the gripper maximum clearance with the surface.

We define three tiers to evaluate the performance of our algorithm on scenes of increasing complexity. 
\begin{itemize}
    \item Tier 0: scenes contain 6 items, either all cups, all bowls, or all utensils, with no stacks in the initial state.
    \item Tier 1: scenes contain 4 items each of cups, bowls, and utensils, and have no stacks in the initial state.
    \item Tier 2: scenes contain 4 items each of cups, bowls, and utensils, but we allow stacks of at most 3 objects in the initial state.
\end{itemize}
For Tier 2, we limit initial stacks to at most 3 objects because of the dimensions of the gripper, as mentioned in Section \ref{stack_policy_subsection}. The number of objects in a stack, and not the actual dimensions of individual dishes, is the main limiting factor for the grasp, because we grasp dishes from the rim. The dishes could actually be much larger and still be graspable as long as the walls are thin enough to allow the gripper to slide over them, and the weight of the dish does not exceed the payload limitations of the gripper itself. We limit ourselves to a small set of known kitchenware objects for consistency in our experiments.

We evaluate the performance of the pull and stack policies against a baseline single-item policy, referred to as ``Random" in Table \ref{tab:physical_experiments}. This policy picks a dish at random, and if the dish is a cup or bowl, it uniformly samples a point on the rim and grasps the dish at that point. If the dish is a utensil, it identifies the grasp point of the utensil as described above and grasps the utensil at that point. This policy is stack-agnostic, so even in Tier 2 when there are stacks present in the initial state, it treats each item in the stack as its own object, and clears the stack by transporting one item at a time. 
\vspace{-2mm}
\subsection{Scene Generation} In order to evaluate our policies, we generate multiple scenes at each tier, and every policy is run once on each scene. To generate each scene, we use the dimensions of the workspace (78cm $\times$ 61cm), and $r_b, r_c, r_u$ for the dimensions of the objects. We randomly sample $x, y$ locations within the scene for each object. If an object intersects with another object, we create a stack of the two objects if the maximum number of intersections has not been exceeded, and resample a position for the object if it has. Tiers 0 and 1 allow no such intersections, whereas Tier 2 allows 4 intersections. For each trial we manually reset the scene to maintain consistency.   
\vspace{-2mm}
\begin{center}
\begin{table*}[t]
\centering
\begin{tabular}{ l  r  c  c  c  c  c } 
\toprule
Tier & \textbf{Policy} & Time (sec) & Objects per Trip & Failures & Time Ratio & \textbf{OpT Ratio} \\ 
\midrule
Tier 0 Cups & \textbf{Random} & 78.2 & 0.8 & 0 & - & - \\
 & \textcolor{blue}{\textbf{Stack}} & 58.5 & 2.0 & 0 & 1.4 & \textcolor{blue}{\textbf{2.6}}  \\
 & \textcolor{VioletRed}{\textbf{Pull}} & 48.8 & 1.6 & 2 & 1.6 & \textcolor{VioletRed}{\textbf{2.1}} \\
 
\midrule
Tier 0 Bowls  & \textbf{Random} & 63.3 & 1.0 & 0 & - & - \\
 & \textcolor{blue}{\textbf{Stack}} & 60.2 & 2.0 & 0 & 1.1 & \textcolor{blue}{\textbf{2.0}} \\ 
 & \textcolor{VioletRed}{\textbf{Pull}} & 41.3 & 1.8 & 0 & 1.5 & \textcolor{VioletRed}{\textbf{1.8}} \\

\midrule
Tier 0 Utensils  & \textbf{Random} & 64.1 & 1.0 & 0 & - & - \\
 & \textcolor{blue}{\textbf{Stack}} & 64.3 & 1.8 & 0 & 1.0 & \textcolor{blue}{\textbf{1.8}} \\
 & \textcolor{VioletRed}{\textbf{Pull}} & 55.3 & 1.8 & 1 & 1.2 & \textcolor{VioletRed}{\textbf{1.8}}  \\ 

\midrule
Tier 1 & \textbf{Random} & 121.3 & 1.0 & 1 & - & - \\
 & \textcolor{blue}{\textbf{Stack}} & 111.3 & 2.0 & 2 & 1.1 & \textcolor{blue}{\textbf{2.0}} \\ 
 & \textcolor{VioletRed}{\textbf{Pull}} & 102.1 & 1.6 & 3 & 1.2 & \textcolor{VioletRed}{\textbf{1.6}} \\ 

 \midrule
Tier 2 & \textbf{Random} & 93.5 & 1.4 & 0 & - & - \\
 & \textcolor{blue}{\textbf{Stack}} & 88.2 & 2.6 & 2 & 1.1 & \textcolor{blue}{\textbf{1.8}} \\ 
 & \textcolor{VioletRed}{\textbf{Pull}} & 84.2 & 2.3 & 3 & 1.1 & \textcolor{VioletRed}{\textbf{1.6}} \\

 \bottomrule
\end{tabular}
\caption{\label{tab:physical_experiments}\textbf{Physical Experiments} We present the total time and number of trips to clear a table for each policy. We found that compared to the baseline policy, the \textcolor{blue}{\textbf{stack}} policy makes at least a \textcolor{blue}{\textbf{1.8x}} improvement in the number of objects grasped per trip (OpT) and the \textcolor{VioletRed}{\textbf{pull}} policy makes at least a \textcolor{VioletRed}{\textbf{1.6x}} improvement.}
\end{table*}
\end{center}

\vspace{-7mm}
\subsection{Evaluation}
We evaluated on 9 scenes at Tier 0 (3 scenes per type of dish), 3 scenes at Tier 1, and 3 scenes at Tier 2. A trial is one execution of one policy on one scene, so we have a total of $(9+3+3)*3 = 45$ trials. For each trial, we record the time in seconds to clear the table, the OpT, and the number of failures. A failure occurs when the robot is unable to move all items to the collection bin, either because of a perception failure that leaves items behind on the workspace or a policy failure that drops a dish off the workspace. We report our results in Table \ref{tab:physical_experiments}.

To evaluate the performance of our policies in more realistic scenario, we present the theoretical improvement in execution time when the bin is placed further away from the workspace, as might be seen in a home or professional kitchen. Given the physical limitation of the UR5 arm length, we simulated the lengthening distance by adding time delays of 3 and 5 seconds in both directions of motion (to and from the collection bin). We find that moving the bin further away causes the stack and pull policies to perform significantly better than the baseline policy because motions to and from the bin are penalized, making policies with fewer total actions perform better. We report these results in Table III in the appendix of the project website.
\vspace{-1mm}

\vspace{1mm}
\section{Discussion} 
Results show that using consolidation and multi-object grasps allows clearing the workspace efficiently, with the pull policy transporting at least 1.6x as many objects per trip, and the stack policy at least 1.8x. A discussion of resulting execution time improvement is in the appendix of the project website.

\section{Limitations and Future Work}
An overhead RGBD camera gives only a clear top view. This affects state estimation and can lead to failures. We assume circular cups and bowls. This makes it easy to compute grasps. For more general dishes, advanced grasp generation methods will be needed. In future work, we will loosen the assumption of starting with singulated objects. We also hope to combine the pull and stack policies into a higher-level policy that can efficiently clear the workspace. 

\section{Acknowledgments}
This research was performed at the AUTOLAB at UC Berkeley in
affiliation with the Berkeley AI Research (BAIR) Lab,
and the CITRIS ``People and Robots" (CPAR) Initiative. The authors were supported in part by donations from Toyota Research
Institute, Bosch, Google, Siemens, and Autodesk and by equipment
grants from PhotoNeo, NVidia, and Intuitive Surgical. Mehmet Dogar was partially supported by an EPSRC Fellowship (EP/V052659).
\vspace{-5mm}
\bibliographystyle{IEEEtran}
{\let\clearpage\relax \vspace{5mm} \bibliography{main} }

\begin{thebibliography}{10}
\providecommand{\url}[1]{#1}
\csname url@rmstyle\endcsname
\providecommand{\newblock}{\relax}
\providecommand{\bibinfo}[2]{#2}
\providecommand\BIBentrySTDinterwordspacing{\spaceskip=0pt\relax}
\providecommand\BIBentryALTinterwordstretchfactor{4}
\providecommand\BIBentryALTinterwordspacing{\spaceskip=\fontdimen2\font plus
\BIBentryALTinterwordstretchfactor\fontdimen3\font minus
  \fontdimen4\font\relax}
\providecommand\BIBforeignlanguage[2]{{%
\expandafter\ifx\csname l@#1\endcsname\relax
\typeout{** WARNING: IEEEtran.bst: No hyphenation pattern has been}%
\typeout{** loaded for the language `#1'. Using the pattern for}%
\typeout{** the default language instead.}%
\else
\language=\csname l@#1\endcsname
\fi
#2}}

\bibitem{DBLP:journals/corr/abs-2110-06192}
A.~X. Lee, C.~Devin, Y.~Zhou, T.~Lampe, K.~Bousmalis, J.~T. Springenberg,
  A.~Byravan, A.~Abdolmaleki, N.~Gileadi, D.~Khosid, C.~Fantacci, J.~E. Chen,
  A.~Raju, R.~Jeong, M.~Neunert, A.~Laurens, S.~Saliceti, F.~Casarini, M.~A.
  Riedmiller, R.~Hadsell, and F.~Nori, ``Beyond pick-and-place: Tackling
  robotic stacking of diverse shapes,'' \emph{CoRR}, 2021.

\bibitem{yamada2009grasp}
T.~Yamada, S.~Yamanaka, M.~Yamada, Y.~Funahashi, and H.~Yamamoto, ``Grasp
  stability analysis of multiple planar objects,'' in \emph{IEEE International
  Conference on Robotics and Biomimetics}, 2009.

\bibitem{9812388}
Y.~Sun, E.~Amatova, and T.~Chen, ``Multi-object grasping - types and
  taxonomy,'' in \emph{IEEE ICRA}, 2022.

\bibitem{9636777}
T.~Chen, A.~Shenoy, A.~Kolinko, S.~Shah, and Y.~Sun, ``Multi-object grasping
  – estimating the number of objects in a robotic grasp,'' in \emph{IEEE
  IROS}, 2021.

\bibitem{Agboh-ISRR-2022}
W.~C. Agboh, J.~Ichnowski, K.~Goldberg, and M.~R. Dogar, ``Multi-object
  grasping in the plane,'' in \emph{ISRR}, 2022.

\bibitem{app12094193}
U.~J. Fernández, S.~Elizondo, N.~Iriarte, R.~Morales, A.~Ortiz, S.~Marichal,
  O.~Ardaiz, and A.~Marzo, ``A multi-object grasp technique for placement of
  objects in virtual reality,'' \emph{Applied Sciences}, 2022.

\bibitem{agboh2022learning}
W.~C. Agboh, S.~Sharma, K.~Srinivas, M.~Parulekar, G.~Datta, T.~Qiu,
  J.~Ichnowski, E.~Solowjow, M.~Dogar, and K.~Goldberg, ``Learning to
  efficiently plan robust frictional multi-object grasps,'' \emph{CoRR}, 2022.

\bibitem{jiang2023multipleobject}
P.~Jiang, J.~Oaki, Y.~Ishihara, and J.~Ooga, ``Multiple-object grasping using a
  multiple-suction-cup vacuum gripper in cluttered scenes,'' 2023.

\bibitem{Nguyen-MOG}
V.~P. Nguyen and W.~T. Chow, ``Wiring-claw gripper for soft-stable picking up
  multiple objects,'' \emph{IEEE Robotics and Automation Letters}, vol.~8,
  no.~7, pp. 3972--3979, 2023.

\bibitem{https://doi.org/10.48550/arxiv.2207.02347}
H.~Huang, L.~Fu, M.~Danielczuk, C.~M. Kim, Z.~Tam, J.~Ichnowski, A.~Angelova,
  B.~Ichter, and K.~Goldberg, ``Mechanical search on shelves with efficient
  stacking and destacking of objects,'' \emph{ISRR}, 2022.

\bibitem{doi:10.1177/0278364912438781}
Y.~Jiang, M.~Lim, C.~Zheng, and A.~Saxena, ``Learning to place new objects in a
  scene,'' \emph{IJRR}, vol.~31, no.~9, pp. 1021--1043, 2012.

\bibitem{Berretty2001OrientingPB}
R.-P. Berretty, K.~Goldberg, M.~H. Overmars, and A.~van~der Stappen,
  ``Orienting parts by inside-out pulling,'' \emph{IEEE ICRA}, 2001.

\bibitem{Huang2017ExactBO}
E.~Huang, A.~Bhatia, B.~Boots, and M.~T. Mason, ``Exact bounds on the contact
  driven motion of a sliding object, with applications to robotic pulling,'' in
  \emph{Robotics: Science and Systems}, 2017.

\bibitem{Satish-RAL-2019}
V.~Satish, J.~Mahler, and K.~Goldberg, ``On-policy dataset synthesis for
  learning robot grasping policies using fully convolutional deep networks,''
  \emph{IEEE Robotics and Automation Letters}, 2019.

\bibitem{8360035}
J.~Mahler, F.~T. Pokorny, S.~Niyaz, and K.~Goldberg, ``Synthesis of
  energy-bounded planar caging grasps using persistent homology,'' \emph{IEEE
  Transactions on Automation Science and Engineering}, 2018.

\bibitem{https://doi.org/10.48550/arxiv.1905.10520}
A.~Mousavian, C.~Eppner, and D.~Fox, ``6-dof graspnet: Variational grasp
  generation for object manipulation,'' \emph{ICCV}, 2019.

\bibitem{DBLP:journals/corr/abs-2002-12738}
M.~Hasan, M.~Warburton, W.~C. Agboh, M.~R. Dogar, M.~Leonetti, H.~Wang,
  F.~Mushtaq, M.~Mon{-}Williams, and A.~G. Cohn, ``Human-like planning for
  reaching in cluttered environments,'' \emph{ICRA}, 2020.

\bibitem{Tirumala-IROS-2022}
S.~Tirumala, T.~Weng, D.~Seita, O.~Kroemer, Z.~Temel, and D.~Held, ``Learning
  to singulate layers of cloth using tactile feedback,'' in \emph{IEEE/RSJ
  IROS}, 2022, pp. 7773--7780.

\end{thebibliography}

\end{document}